\documentclass[conference]{IEEEtran}
\IEEEoverridecommandlockouts
\usepackage{cite}
\usepackage{amsmath,amssymb,amsfonts}
\usepackage{algorithmic}
\usepackage{graphicx}
\usepackage{textcomp}
\usepackage{xcolor}
\def\BibTeX{{\rm B\kern-.05em{\sc i\kern-.025em b}\kern-.08em
    T\kern-.1667em\lower.7ex\hbox{E}\kern-.125emX}}
\begin{document}

\title{Hybrid Transformer-RNN Architecture for Household Occupancy Detection Using Low-Resolution Smart Meter Data
% {\footnotesize \textsuperscript{*}Note: Sub-titles are not captured in Xplore and should not be used}
\thanks{This work was supported in part by the Australian Research Council (ARC) Discovery Early Career Researcher Award (DECRA) under Grant DE230100046.}
}

\author{\IEEEauthorblockN{Xinyu Liang$^{1}$, Hao Wang$^{1,2*}$}
\IEEEauthorblockA{
    $^{1}$\textit{Department of Data Science and AI, Faculty of IT, Monash University, Melbourne, VIC 3800, Australia} \\
    $^{2}$\textit{Monash Energy Institute, Monash University, Melbourne, VIC 3800, Australia} \\
    Emails: adamliang42@gmail.com, hao.wang2@monash.edu}
\thanks{*Corresponding author: Hao Wang.}
}

\maketitle

\begin{abstract}
Residential occupancy detection has become an enabling technology in today's urbanized world for various smart home applications, such as building automation, energy management, and improved security and comfort. 
Digitalization of the energy system provides smart meter data that can be used for occupancy detection in a non-intrusive manner without causing concerns regarding privacy and data security. In particular, deep learning techniques make it possible to infer occupancy from low-resolution smart meter data, such that the need for accurate occupancy detection with privacy preservation can be achieved. Our work is thus motivated to develop a privacy-aware and effective model for residential occupancy detection in contemporary living environments. 
Our model aims to leverage the advantages of both recurrent neural networks (RNNs), which are adept at capturing local temporal dependencies, and transformers, which are effective at handling global temporal dependencies. 
Our designed hybrid transformer-RNN model detects residential occupancy using hourly smart meter data, achieving an accuracy of nearly 92\% across households with diverse profiles. 
We validate the effectiveness of our method using a publicly accessible dataset and demonstrate its performance by comparing it with state-of-the-art models, including attention-based occupancy detection methods.
\end{abstract}

\begin{IEEEkeywords}
Occupancy detection, smart meter data, deep learning, transformer, recurrent neural network (RNN)
\end{IEEEkeywords}

\section{Introduction}
The significance of residential occupancy detection has increased substantially, primarily driven by global urbanization and concurrent population growth in recent years~\cite{barbour2019planning}. Accurately determining patterns of occupancy holds utmost importance, as it can improve self-awareness of occupancy patterns for residents and enable various business opportunities for utility companies and building managers, including energy saving, thermal comfort control, and route optimization for work activities and deliveries~\cite{brooks2014energy,ohsugi2018delivery}. Consequently, accurate occupancy detection yields a range of benefits, including economic advantages, positive environmental impacts, and enhanced security and comfort for residents. 

% A considerable amount of research has been conducted on residential occupancy detection, with one common approach involving the installation of supplementary cameras or sensors, such as thermal imaging cameras or motion sensors \cite{chidurala2021occupancy, savazzi2019occupancy, gu2019dynamic, maaspuro2020low}. These sensors collect data that is then utilized to develop occupancy detection systems or algorithms. Although numerous studies demonstrate the high accuracy of these methods, there are several drawbacks associated with using additional sensors for residential occupancy detection. Firstly, the installation of multiple sensors throughout a residence can be a significant undertaking, as it may involve invasive modifications to the home and often necessitates regular maintenance and calibration, which can be both time-consuming and financially challenging for homeowners. Another concern is the potential invasion of privacy, as occupants may feel uncomfortable with constant monitoring, even if the primary objective is to improve energy efficiency or security. Moreover, incorporating numerous sensors into an occupancy detection system introduces complexity due to the need for advanced algorithms and software. These drawbacks make this type of method hard to scale and can potentially cause privacy concerns.

A large body of research has been conducted on residential occupancy detection. Many existing studies adopted one common approach involving the installation of supplementary cameras or sensors, such as thermal imaging cameras or motion sensors \cite{chidurala2021occupancy, savazzi2019occupancy, gu2019dynamic, maaspuro2020low}. Though these methods could achieve high accuracy, they may cause significant concerns or pose new challenges. For example, installing sensors is intrusive, requires regular maintenance, and can be costly. Furthermore, constant monitoring can invade occupants' privacy, raising ethical concerns. Lastly, the integration of multiple sensors increases system complexity, which in turn requires sophisticated algorithms and software. Thus, these methods may face scalability issues and raise concerns regarding privacy risks.

% Compared to camera or motion-sensing based methods, smart meters are already integrated into the utility infrastructure so that they can effectively reduce installation costs and maintenance responsibilities for homeowners. For the same reason, smart meter data can be more readily incorporated with other systems, such as energy management or home automation platforms, as it seamlessly connects with existing utility and appliance systems. Furthermore, the data collected from smart meters is inherently less sensitive than the information captured by motion or camera-based systems inside the home. Recent studies \cite{gao2018occupancy,ohsugi2018delivery,yilmaz2021avoiding,feng2020deep} utilizing high-resolution smart meter data and machine-learning techniques have achieved accuracy levels comparable to dedicated sensor systems. However, the detailed information found in high-resolution smart meter data can reveal occupants' habits and behavior, potentially raising privacy concerns. Additionally, frequent sampling is more likely to cause data transmission, storage, and processing issues on a large scale. 

To address the aforementioned issues, researchers have been studying non-intrusive approaches. Compared to camera or motion-sensing-based methods, smart meters, widely installed as a part of utility infrastructure, can provide an alternative and cost-effective approach to occupancy detection. The energy consumption data monitored by smart meters can be easily integrated with energy management or home automation systems, providing an inherent advantage over the camera or motion-sensing methods as there is no need for extra installation and maintenance. Recent studies using high-resolution smart meter data \cite{gao2018occupancy,ohsugi2018delivery,yilmaz2021avoiding,feng2020deep} have achieved comparable accuracy levels to other methods using sensors. However, privacy concerns still arise due to the detailed energy consumption information revealing occupants' habits. Additionally, frequent sampling also cause scalability issues to data transmission, storage, and processing. Thus, it is crucial to avoid using high-resolution data but using low-resolution data, which can preserve privacy.

Low-resolution smart meter data occupancy detection is proposed to further resolve privacy and scalability limitations. However, the reduced information content in low-resolution smart meter data makes occupancy detection tasks quite challenging to achieve high accuracy equivalent to other methods using high-resolution data. Deep learning techniques provide promising solutions to improve the detection accuracy using low-resolution data. For example, Hisashi et al. \cite{oshima2022occupancy} proposed a deep learning-based method to estimate residential occupancy status. Their proposed method included manual feature extraction to derive statistical features from time-series sequences and subsequently process the extracted data through a bi-directional long short term memory (BI-LSTM) network-a commonly used recurrent neural network (RNN), with the attention mechanism. Moreover, their method trained separate models for individual households, leading to generalization problems. According to \cite{wang2021identifying,tang2022machine}, households with diverse socioeconomic characteristics exhibit different energy consumption profiles. An effective occupancy detection method should overcome generalization limitations and be applied to a broad groups of households with diverse socioeconomic backgrounds and lifestyles.

% In order to mitigate the shortcomings of current methods, we have devised an innovative deep learning-based approach for residential occupancy detection, with the goal of attaining state-of-the-art performance in the context of low-resolution smart meter occupancy detection. We evaluated our methodology on the most extensive publicly accessible dataset, and the results showcase enhanced performance across a diverse range of households collectively. Our contributions can be summarized as follows:
To overcome the limitations of existing methods, we are motivated to design a new deep-learning-based approach for residential occupancy detection using low-resolution smart meter data while achieving considerably high accuracy. Specifically, we employ a hybrid transformer-Bi-LSTM architecture that enables processing of raw smart meter data without the need for manual feature extraction. In addition, our model is designed to be applicable to various households rather than being trained for each individual household separately. The goal of our work is to achieve state-of-the-art performance for occupancy detection using low-resolution smart meter. To evaluate the effectiveness of our model, we conduct experiments on the most comprehensive publicly accessible dataset \cite{kleiminger2015household}. The results show that our model improves occupancy detection performance across households compared to baseline methods. The contributions of our work are summarized as follows.
\begin{itemize}
    \item Our work presents a novel model by combining RNNs and transformers to effectively model temporal dependencies in low-resolution smart meter data. By leveraging RNNs' capability in sequential processing of short to medium-term  dependencies and transformers' self-attention for long-range dependencies, we enhance the performance and accuracy of occupancy detection.
    \item Our work explores various transformer-RNN hybrid models by thoroughly examining the fusion of these architectures in different arrangements. Through our investigation, we find an optimal combination: the concatenation of Bi-LSTM and transformers. Our design leverages the temporal modeling of Bi-LSTM and transformers' self-attention mechanism, shedding light on the effective construction of such hybrid models for similar tasks.
    % \item We used combined data from various households to evaluate our model's generalization across different scenarios. Performance was then rigorously tested using multiple metrics and validated with a 10-fold cross-validation method. The results consistently showed our model outperforms existing methods in occupancy detection across all metrics, confirming its robustness across a diverse range of households.
    \item We compare our model to different models in residential occupancy detection using a comprehensive benchmarking framework, including various performance metrics and cross-validation, based on a real-world household dataset. Our findings demonstrate that the fusion of transformers and Bi-LSTM models through a concatenation operation consistently outperforms other baseline models in terms of a comprehensive set of performance metrics.
\end{itemize}

The remainder of this paper is organized as follows. Section \ref{sec:method} presents the problem formulation and the hybrid transformer-RNN model for occupancy detection. Section \ref{sec:benchmark} introduces the benchmark models and evaluation metrics. Section \ref{sec:result} discusses numerical results, and Section \ref{sec:conclu} concludes this paper.

% \begin{itemize}
%     \item We integrated RNNs and Transformers to facilitate comprehensive modeling of temporal dependencies within low-resolution smart meter data at various scales. Our model employs the sequential processing capabilities of RNNs to capture short to medium-term dependencies while leveraging the non-sequential processing approach of Transformers to identify long-range dependencies through self-attention mechanisms. This combination results in enhanced performance and increased accuracy in occupancy detection classification.
%     \item Rather than relying on manual temporal feature extraction, our model utilizes raw smart meter data, delegating feature extraction to learning-based methods. This approach enhances scalability across various households and, as indicated by the evaluation results, improves adaptability to changes in data characteristics and distributions among different households.
%     \item Rather than training distinct models for each individual household, we chose to develop a single model using smart meter data gathered from a variety of households. This strategy facilitates the determination of the model's applicability across a broad spectrum of household situations, thereby improving the model's generalizability and adaptability. Furthermore, this approach has the potential to simplify the model management process, as only one model needs to be maintained, and may yield more consistent performance in residential occupancy detection tasks, even among households with diverse characteristics.
% \end{itemize}
\section{Methodology}\label{sec:method}
\begin{figure}[t]
% \centerline{\includegraphics[width=\columnwidth]{Figs/model_detailed.pdf}}
\centerline{\includegraphics[width=0.9\columnwidth]{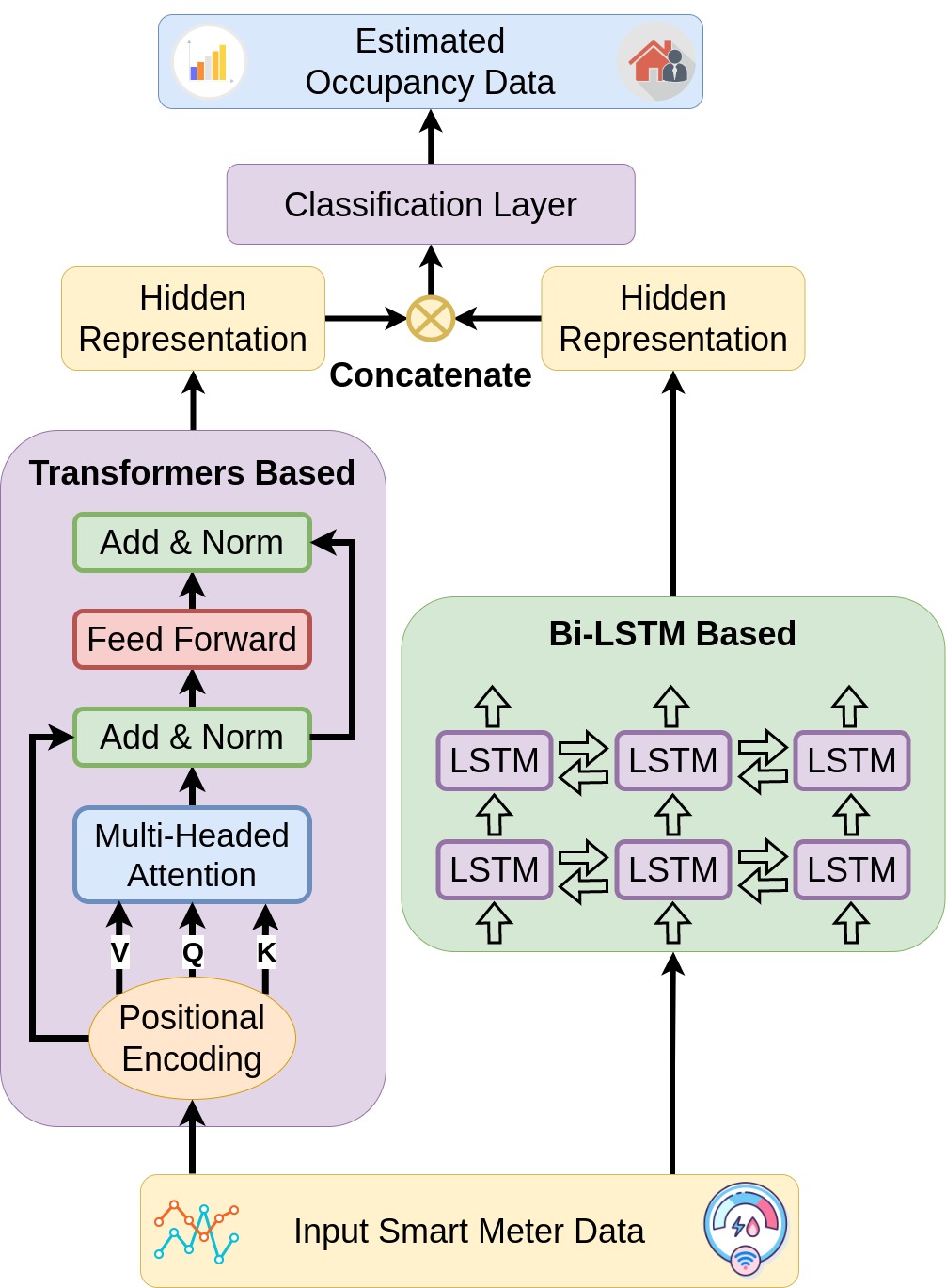}}
\caption{\centering{Hybrid Transformer-RNN Model for occupancy detection.}}
\label{low_level_fig}
\vspace{-3mm}
\end{figure}

We present a novel model aiming to enhance the performance of occupancy detection by leveraging a hybrid transformer-Bi-LSTM architecture on low-resolution smart meter data. As depicted in Figure \ref{low_level_fig}, the smart meter data is directly fed into both the transformer-based and Bi-LSTM-based feature extractors. Then the extracted features from these two components are concatenated to form a comprehensive feature set. This amalgamated feature set is inputted into the classification layer, which discerns the presence or absence of the occupants in each time step, e.g., each hour.
% Such additional data sources are frequently linked to elevated installation expenses and privacy-related concerns. 
In the following, we will present the problem formulation and our model in detail.

% \begin{figure}[ht]
% \centerline{\includegraphics[width=\columnwidth]{Figs/occupance_detection.png}}
% \caption{\centering{Smart meter Data for occupancy estimation.}}
% \label{high_level_fig}
% \vspace{-3mm}
% \end{figure}

\subsection{Problem Formulation}
Given a dataset denoted by $\mathcal{D} = {(\mathbf{X}_i, \mathbf{y}_i)}_{i=1}^{N}$, containing $N$ samples, and $i=1,...,N$. Each sample consists of a time series of smart meter readings denoted as $\mathbf{X}_i \in \mathbb{R}^{T \times F}$ with a length of $T$ and $F$ features. An occupancy status label is introduced as $\mathbf{y}_i \in \{0, 1\}^T$, where $1$ indicates the presence of occupants and $0$ indicates the absence. Each individual time-series sequence $\mathbf{X}_i$ can be decomposed into $(\mathbf{x}_{i,1}, \mathbf{x}_{i,2}, \dots, \mathbf{x}_{i,T})$, where $\mathbf{x}_{i,t} \in \mathbb{R}^F$ serves as the $F-$dimensional feature vector at a given time step $t$ for the $i$-th sample. A similar decomposition is applicable to the occupancy status labels, manifesting as $\mathbf{y}_i = (y_{i,1}, y_{i,2}, \dots, y_{i,T})$, where $y_{i,t}$ signifies the occupancy status at time step $t$ for the $i$-th sample.

The time-series sequences $\mathbf{X}_i$ contain valuable information regarding household energy consumption patterns and occupancy-driven behaviors. This information can be harnessed to estimate occupancy status. The aim of our model is to design a deep learning model $f(\mathbf{X}; \boldsymbol{\theta})$, parameterized by $\boldsymbol{\theta}$. The model is capable of predicting the occupancy status $\hat{\mathbf{y}}$ from the input smart meter data $\mathbf{X}$, i.e., $\hat{\mathbf{y}} = f(\mathbf{X}; \boldsymbol{\theta})$, such that $\hat{\mathbf{y}}$ can accurately predict $\mathbf{y}$.

The objective of the occupancy prediction is to minimize the loss function $L$, which is the binary cross entropy between the predicted occupancy status $\hat{\mathbf{y}}$ and the actual occupancy status $\mathbf{y}$:
\begin{equation}
L(\mathcal{D},\boldsymbol{\theta}) = \frac{1}{N} \sum_{i=1}^{N} \ell(f(\mathbf{X}_{i}; \boldsymbol{\theta}), \mathbf{y}_{i}),
\end{equation}
in which $\ell$ denotes the binary cross-entropy loss for time-series classification. By optimizing the model parameters $\boldsymbol{\theta}$, the deep learning model is primed to accurately classify the occupancy status at each time step.
% , while also demonstrating robust generalization to unseen data and adaptability to a diverse array of household scenarios.

% \subsection{Bidirectional Long Short Term Memory (Bi-LSTM)}
\subsection{Feature Extraction Utilizing Bi-LSTM}
% As shown in Figure \ref{low_level_fig} we integrated a Bi-LSTM Based feature extractor inside our model.
% Long Short-Term Memory (LSTM) networks \cite{hochreiter1997long} are a specific type of Recurrent Neural Networks (RNN) that introduce memory cells and gating mechanisms that enable the network to learn and retain information over longer sequences effectively. Consequently, LSTM is capable of extracting long and complex temporal features dependencies in smart meter data for residential occupancy detection.
Due to the inherently temporal nature of smart meter data, RNNs, such as Long Short-Term Memory (LSTM) networks \cite{hochreiter1997long}, are a suitable architecture for feature extraction. Equipped with memory cells and gating mechanisms, LSTM networks are capable of effectively learning and retaining information over lengthy sequences. This results in an enhanced ability to extract long and intricate temporal feature dependencies, playing a crucial role in detecting occupancy accurately. In particular, the Bi-LSTM extends the capabilities of LSTM by processing the input sequence in both forward and backward directions. This allows the model to capture information from both past electricity consumption information and future usage data, providing a more comprehensive understanding of the context. A Bi-LSTM consists of two LSTM layers: a forward layer and a backward layer. For any given smart meter data $\textbf{X} = (\textbf{x}_1, \textbf{x}_2, ..., \textbf{x}_T)$, the forward LSTM computes $\overrightarrow{\mathbf{H}}_{rnn} = (\overrightarrow{\mathbf{h}}_1, \overrightarrow{\mathbf{h}}_2, ..., \overrightarrow{\mathbf{h}}_T)$, in which $\overrightarrow{\mathbf{h}}_t$ is the hidden state for each time step. Similarly, the backward LSTM computes the hidden states $\overleftarrow{\mathbf{H}}_{rnn} = (\overleftarrow{\mathbf{h}}_1, \overleftarrow{\mathbf{h}}_2, ..., \overleftarrow{\mathbf{h}}_T)$, in which $\overleftarrow{\mathbf{h}}_t$ is is the hidden state for each time step. The final extracted temporal features at each time step are then obtained by concatenating the forward and backward hidden states as $\mathbf{H}_{rnn} = \text{Concat}(\overrightarrow{\mathbf{H}}_{rnn}; \overleftarrow{\mathbf{H}}_{rnn})$.

% \begin{figure}[t]
% % \centerline{\includegraphics[width=\columnwidth]{Figs/model_detailed.pdf}}
% \centerline{\includegraphics[width=\columnwidth]{Figs/model_detailed.jpg}}
% \caption{\centering{Hybrid Transformer-RNN Model for occupancy detection.}}
% \label{low_level_fig}
% \vspace{-3mm}
% \end{figure}

\subsection{Exploiting Transformers Encoder for Feature Extraction}
While the Bi-LSTM can significantly contribute to feature extraction, relying solely on it may not fully encapsulate the complex dynamics of smart meter data. Specifically, one inherent constraint of LSTMs is their sequential processing nature, which assumes a certain chronological order in data. While this characteristic is advantageous in capturing local temporal dependencies, it may overlook the non-sequential patterns and long-range dependencies that could exist within the household electricity consumption behavior. Hence, to improve the understanding of these dynamics, we incorporate the transformer \cite{vaswani2017attention} into our feature extraction process. Unlike LSTM, transformers do not necessitate a sequential processing approach but attend to different parts of the sequence regardless of their position. This is made possible through the employment of self-attention mechanisms, making transformers highly effective across diverse domains. For the purpose of feature extraction from smart meter data for occupancy detection, we utilize solely the encoder part of the transformer architecture.
% The Transformer architecture was introduced by Vaswani et al. \cite{vaswani2017attention} to process sequence data. It employs unique self-attention mechanisms and has been successfully applied in various domains, including natural language processing and time series analysis. The architecture of the Transformer includes an encoder and a decoder. However, we only require the encoder part to extract temporal features from smart meter data for occupancy detection. 

Given the input time series of smart meter data $\mathbf{X}$, we apply positional encodings to generate a modified data matrix $\mathbf{X'} = \mathbf{X} + \text{PE}$. 
% This operation infuses the model with an intrinsic awareness of the temporal structure inherent in the sequence, thereby enabling the model to accommodate the sequential nature of time series data. 
% The positional encodings (PE) are presented as
% \begin{align}
% \text{PE}{(p, 2m)} &= \sin\left(\frac{p}{10000^{2m/F}}\right), \\
% \text{PE}{(p, 2m + 1)} &= \cos\left(\frac{p}{10000^{2m/F}}\right), 
% \end{align}
% where $p$ is the position, $m$ is the index into the dimensions of the smart meter features (divided by 2 because the encoding function alternates between sine and cosine). This operation infuses the model with an intrinsic awareness of the temporal structure inherent in the smart meter data to avoid the limitation that the self-attention mechanism itself does not have any awareness of the electricity usage order.

The positional encodings (PE) are presented as
\begin{align}
\text{PE}{(p, 2m)} &= \sin\left(\frac{p}{10000^{2m/F}}\right), \\
\text{PE}{(p, 2m + 1)} &= \cos\left(\frac{p}{10000^{2m/F}}\right), 
\end{align}
where $p$ is the position of the time step, and $m$ is the index of the smart meter features (divided by 2 because the encoding function alternates between sine and cosine). This operation infuses the model with an intrinsic awareness of the temporal structure in the smart meter data. This overcomes the limitation that the self-attention mechanism itself does not have any awareness of the electricity usage order.

Following the application of positional encodings, the model proceeds with the multi-head self-attention mechanism. This component enables the model to focus on different parts of the input sequence for every attention head, thereby capturing a richer set of dependencies. With a total of $U$ heads, for each attention head, $\mathbf{X}'$ is linearly transformed to generate corresponding queries ($\mathbf{Q}$), keys ($\mathbf{K}$), and values ($\mathbf{V}$). This is captured in the following equations for the $u$-th head:
\begin{equation}
\mathbf{Q}_u= 
\mathbf{X}'\mathbf{W}^{Q}_{u},
\quad\mathbf{K}_u= \mathbf{X}'\mathbf{W}^{K}_{u},
\quad\mathbf{V}_u= \mathbf{X}'\mathbf{W}^{V}_{u},
\end{equation}
where $\mathbf{W}^{Q}_{u}$, $\mathbf{W}^{K}_{u}$, and $\mathbf{W}^{V}_{u}$ are learned weight matrices. After obtaining the projected query, key, and value of dimension $d_k$ for each head, we compute the value of the head using scaled dot-product attention as
\begin{align}
\text{head}_u = \text{Att}(\mathbf{Q}_u, \mathbf{K}_u, \mathbf{V}_u) = \text{softmax}\left(\frac{\mathbf{Q}_u\mathbf{K}_u^\top}{\sqrt{d_k}}\right)\mathbf{V}_u.
% \text{head}_u = \text{Attention}(\mathbf{Q}_u, \mathbf{K}_u, \mathbf{V}_u),
\end{align}

The outputs will then be concatenated and linearly transformed using parameter matrices $\mathbf{W}^O$ to yield the final output of multi-head self-attention shown as
\begin{equation}
    \text{MultiHead}({\mathbf{X}'}) = \text{Concat}(\text{head}_1,...,\text{head}_U)\mathbf{W}^O.
\end{equation}
After the multi-head self-attention operation, a residual connection is implemented. This technique is prevalent in deep learning to help mitigate the problem of vanishing gradients during training. The output of the multi-head self-attention is added directly to the encoded input. Following the residual connection, layer normalization \cite{ba2016layer} is performed to stabilize the learning dynamics and expedite the training process to achieve $\mathbf{X}''$ shown as
\begin{equation}
\mathbf{X}'' = \text{LayerNorm}(\mathbf{X}' + \text{MultiHead}({\mathbf{X}'})).
\end{equation}
which will then undergoes an additional transformation by a position-wise feed-forward network (FFN), enabling the model to further capture and learn from the relationships among the input data, transforming the abstract representations from the multi-head self-attention mechanism into higher-level features that can more effectively inform the occupancy status predictions. The FFN operates as follows
\begin{equation}
\text{FFN}(\mathbf{X''}) = \text{ReLU}(\mathbf{X}''\mathbf{W}_1 + \mathbf{b}_1) \mathbf{W}_2 + \mathbf{b}_2,
\end{equation}
where $\mathbf{W}_1$, $\mathbf{b}_1$, $\mathbf{W}_2$, and $\mathbf{b}_2$ are learnable parameters of the feed-forward network. The ReLU (Rectified Linear Unit) function is applied as a non-linear activation function.

Following the feed-forward network, a second residual connection and layer normalization process is implemented to produce the final feature representation $\mathbf{H}_{trans}$ of the transformer network, presented as
\begin{equation}
\mathbf{H}_{trans} = \text{LayerNorm}(\mathbf{X}'' + \text{FFN}(\mathbf{X}'')).
\end{equation}
% This combination of residual connection and layer normalization again aids in facilitating training and enhancing the performance of the deep learning model.

\subsection{Occupancy Detection via Concatenated Feature Representation}
% Transformers, known for their prowess in sequence processing tasks, achieve this through a unique non-sequential architecture based on attention mechanisms. Unlike Bi-LSTM's sequential processing. This difference allows Transformers to dynamically assign importance to data points in smart meter data, enhancing understanding of long-term dependencies. However, despite positional encoding, inherent limitations like the lack of temporal bias and local sensitivity persist. To mitigate these, we propose a hybrid approach that combines the sequential Bi-LSTM and non-sequential Transformers, effectively encoding short and long-term features. This enhanced encoding enables more comprehensive insights into occupancy behavior across various households. 
The strengths and limitations of Bi-LSTM and transformers show a noteworthy reciprocal relationship when handling smart meter data. Bi-LSTM with its strong ability in capturing local temporal dependencies and displaying both temporal bias and local sensitivity well compensates for these inherent limitations of transformers. Conversely, transformers equipped with their unique non-sequential architecture can capture long-term dependencies.Therefore, a hybrid architecture that combines Bi-LSTM and transformers can harness the strengths of both while simultaneously mitigating their weaknesses, leading to a more robust and effective model.
In pursuit of this goal, we generate a consolidated hidden representation $\mathbf{H} = \text{Concat}(\mathbf{H}_{rnn}, \mathbf{H}_{trans})$, which concentrates the feature representations obtained from the transformer-based and Bi-LSTM-based components, respectively.
% $\mathbf{H}_{rnn}$ and $\mathbf{H}_{trans}$ formulated as:
% \begin{equation}
%     \mathbf{H} = \text{Concat}(\mathbf{H}_{rnn}, \mathbf{H}_{trans})
% \end{equation}

This hidden representation is then passed through a classification layer to generate estimations of occupancy status for each time step. The classification layer is a dense layer with a sigmoid activation function, which maps each time step of the sequence to a probability between 0 and 1. This probability indicates the model's confidence of occupancy at that particular time step. This process is represented as
\begin{equation}
\hat{\mathbf{y}} = \sigma(\mathbf{H}\mathbf{W}_{c} + \mathbf{b}_{c}),
\end{equation}
where $\mathbf{W}_{c}$ and $\mathbf{b}_{c}$ are the weight matrix and bias vector of the classification layer, respectively, and $\sigma$ denotes the sigmoid activation function.

As our objective is to minimize the binary cross-entropy loss between the predicted and actual occupancy statuses, the loss is computed as
\begin{equation}
\ell(\hat{\mathbf{y}}, \mathbf{y}) = -\frac{1}{T} \sum_{t=1}^{T} \left[ y_{t} \log(\hat{y}_{t}) + (1 - y_{t}) \log(1 - \hat{y}_{t}) \right].
\end{equation}

\section{Benchmarks and Performance Evaluation}\label{sec:benchmark}

% In order to thoroughly compare our proposed Hybrid Transformer-RNN model with standard models, we carried out detailed experiments using the widely available ECO dataset \cite{kleiminger2015household,beckel2014eco}. 
This section introduces the benchmark models we choose for comparisons. We also provide details of the metrics we will use to measure the performance of our model and the cross-validation techniques. 

\subsection{Benchmarks}
In this paper, we evaluate different variants of transformers-RNN hybrid models, as well as deep learning models including attention mechanisms from previous studies on occupancy detection. We aim to provide a robust and comprehensive evaluation of the state-of-the-art in occupancy prediction methods, while also exploring the potential advantages of novel combinations of transformers and Bi-LSTM networks. Below, we provide a brief description for each of these examined models.
\begin{itemize}
    \item \textbf{Bi-LSTM + Transformers:} This model first processes the input data with a Bi-LSTM and then feeds the output into a transformer. This sequential combination allows the transformer to build upon the temporal dependencies recognized by the Bi-LSTM.
    \item \textbf{Transformers + Bi-LSTM:} This model reverses the order of the previous combination. The input data is first processed by a transformer, and the output is then fed into a Bi-LSTM. This configuration enables the Bi-LSTM to refine the global patterns identified by the transformer.
    \item \textbf{Bi-LSTM + Attention:} Proposed in a previous study \cite{oshima2022occupancy}, this model combines a Bi-LSTM with an attention mechanism, processing the input data with the Bi-LSTM and using the attention mechanism to weigh the importance of different time steps in the output. We evaluate this model using both original and feature-extracted data, with the latter reflecting techniques from the same previous study where manual feature extraction is performed to potentially enhance model performance.
\end{itemize}

\subsection{Evaluation Metrics}
By evaluating the performance of our occupancy detection models, we utilize a diverse set of metrics, each providing distinct perspectives on the functionality of the models. The following offers a detailed description of these metrics.
\begin{itemize}
    \item \textbf{Accuracy}: The ratio of correct to total occupancy predictions. Note that imbalanced occupancy rates may distort this measure and we need to examine other metrics below.
    \item \textbf{Precision}: The fraction of true positive instances (correctly predicted presence) out of all predicted as presence.
    \item \textbf{Recall}: The ratio of true positive instances to all real occupied instances.
    \item \textbf{F1-Score}: Harmonic mean of precision and recall, providing a balanced performance measure.
    \item \textbf{ROC AUC}: It represents the trade-off between recall and the false positive rate. Scores range from 0.5 (random guessing) to 1 (perfect model).
\end{itemize}

Complementing the metrics previously discussed, we employed 10-fold Cross-Validation to enhance the robustness of our occupancy detection model. This technique partitions the original occupancy data into ten equally sized subsets. One of these subsets serves as the validation data for model testing, while the remaining subsets form the training data. This procedure is reiterated ten times, with each time using a different subset as the validation data. The outcomes of these iterations are then combined to yield a unified estimation of model performance. This method assists in countering the risk of overfitting and augments our model's performance estimation on unseen data.
\section{Numerical Results And Analysis}\label{sec:result}

\subsection{Data Description}
The Electricity Consumption \& Occupancy (ECO) dataset \cite{kleiminger2015household} is the largest publicly available dataset for non-intrusive load monitoring and occupancy detection studies to our best knowledge. This comprehensive dataset, gathered over eight months from six Swiss households, provides aggregate and appliance-specific power consumption data at a 1Hz frequency. Each data point includes current, voltage, and phase shift information for each of the household's three electrical phases. Additionally, occupancy information is recorded and obtained through manual labeling via tablets or passive infrared sensors. 

\begin{table}[b]
\centering
\caption{Number of Days and Occupancy Rate for Each Household After Data Preprocessing}
\begin{tabular}{|c|c|c|}
\hline
\textbf{Household Number} & \textbf{Number of Days} & \textbf{Occupancy Ratio} \\
\hline
1 & 85 & 0.7715 \\
\hline
2 & 126 & 0.7100 \\
\hline
3 & 78 & 0.7382 \\
\hline
4 & 86 & 0.9205 \\
\hline
5 & 74 & 0.8868 \\
\hline
\textbf{Total} & 449 & 0.7960 \\
\hline
\end{tabular}
\label{tab:preprocessing_data}
\end{table}

\subsection{Data Preprocessing}
As our study focuses solely on smart meter data, we utilize the aggregated information on current, voltage, and phase shifts to match it with occupancy information based on date and household accordingly. Since our study aims to detect occupancy using low-resolution smart meter data, we resample both the smart meter and occupancy data to a lower resolution to obtain a dataset with one-hour intervals. For the smart meter data, we calculate the average of 3600 measurements within each hour. For the occupancy data, we assess 3600 occupancy statuses within each hour and select the most frequently occurring status. At last, we have compiled a dataset, spanning 449 days from five distinct households with detailed information shown in Table \ref{tab:preprocessing_data}.

% \begin{figure}[ht]
% \centerline{\includegraphics[width=\columnwidth]{Figs/occupancy_result_visualization.pdf}}
% \caption{\centering{Box Plot for 10-fold Cross Validation Result Metric Comparison Across Different Methods}}
% \label{occupancy_result_visualization}
% \vspace{-3mm}
% \end{figure}

\begin{table*}[ht]
\caption{Model evaluation results}
\centering
\begin{tabular}{|c|c|c|c|c|c|}
\hline
\textbf{Model} & \textbf{Accuracy} & \textbf{Precision} & \textbf{Recall} & \textbf{F1 score} & \textbf{ROC AUC} \\
\hline
Bi-LSTM \& Transformers Concatenation with Original Data & \textbf{0.9166} & \textbf{0.9323} & 0.9623 & \textbf{0.9470} & \textbf{0.9331} \\
Bi-LSTM + Transformers with Original Data & 0.9119 & 0.9287 & 0.9601 & 0.9441 & 0.9297 \\
Transformers + Bi-LSTM with Original Data & 0.9065 & 0.9202 & \textbf{0.9628} & 0.9410 & 0.9285 \\
Bi-LSTM + Attention with Original Data & 0.8795 & 0.9075 & 0.9404 & 0.9236 & 0.9109 \\
Bi-LSTM + Attention with Feature Extracted Data & 0.8785 & 0.9021 & 0.9505 & 0.9256 & 0.8971 \\
\hline
\end{tabular}
\label{tab:results}
\end{table*}

\begin{figure}[ht]
\centerline{\includegraphics[width=\columnwidth]{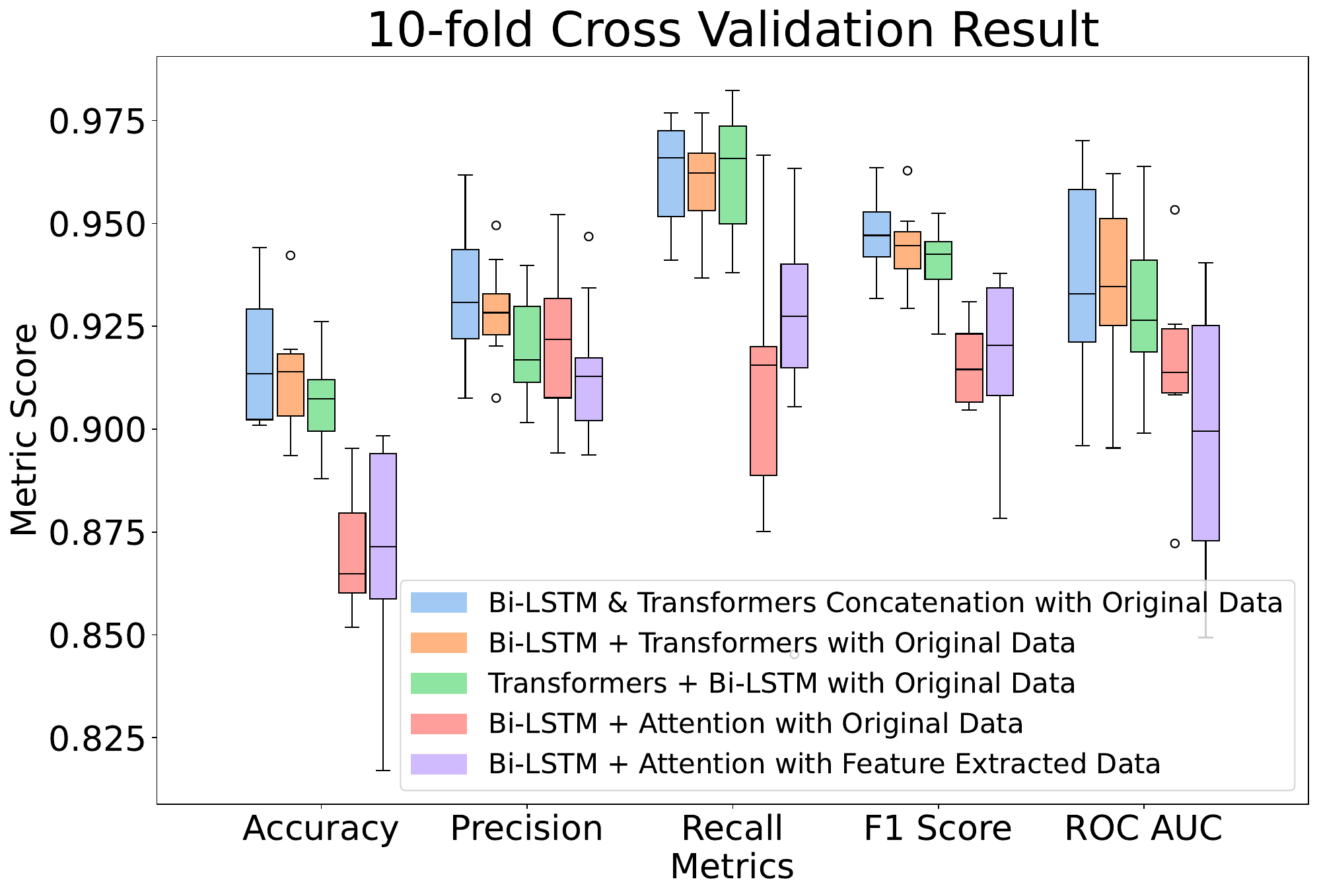}}
\caption{\centering{Box Plot for 10-fold Cross Validation Result Metric Comparison Across Different Methods.}}
\label{10fold_visualization}
\vspace{-3mm}
\end{figure}

\subsection{Results Analysis}

% This section presents a sample of visualized occupancy detection results shown in Figure \ref{occupancy_result_visualization}, and model evaluation results shown in Table \ref{tab:results} and Figure \ref{10fold_visualization}. Following we offer a comparative analysis of different occupancy detection models:

This section presents the model evaluation results shown in Table \ref{tab:results} and Figure \ref{10fold_visualization}. In the following, we offer a comparative analysis of different occupancy detection models.

\subsubsection{Our Bi-LSTM and Transformers Concatenation Model} Our model achieves an accuracy of approximately 92\% in residential occupancy detection, surpassing all benchmark models. The Bi-LSTM and transformers concatenation model has shown a remarkable performance in occupancy detection utilizing original smart meter data (without manual feature extraction), demonstrated by numerical results shown in Table \ref{tab:results} and the corresponding box plot for 10-fold cross-validation shown in Figure \ref{10fold_visualization}. Our model achieves the highest accuracy of 0.9166, indicating an exceptional proficiency in accurately detecting occupancy status from low-resolution smart meter data. 
% Its precision score of 0.9323 emphasizes the model's capability to minimize false-positive detections, a crucial consideration in preventing undue energy expenditure resulting from erroneous occupancy predictions. 
Furthermore, the model achieves the highest F1 score and ROC AUC score, showcasing its ability to strike a balance between precision and recall, effectively managing potential class imbalances in the data, and demonstrating robust discriminative power between different occupancy states. The numerical results not only establish the model's robustness and reliability but are also supported by the box plot distribution depicted in Figure \ref{tab:results}, which shows consistent performance across all evaluation metrics. This reinforces the notion that the hybrid model effectively leverages the strengths of both the Bi-LSTM and transformers.

\subsubsection{Effectiveness of Transformer-RNN Hybrid Models and The Impact of Different Integration Approaches} Among all the Transformer-RNN hybrid models, despite the Bi-LSTM and transformers concatenation version,  the Bi-LSTM + transformers and transformers + Bi-LSTM versions have also exhibited significant performance enhancements over previous Bi-LSTM + attention models for occupancy detection using original smart meter data. The method of integration of transformers and Bi-LSTM models seems to play a vital role based on the effectiveness of these hybrid models, as demonstrated by our numerical results in Table \ref{tab:results} and the corresponding box plots in Figure \ref{10fold_visualization}. The Bi-LSTM + transformers model, which initially processes data via the Bi-LSTM before forwarding it to the transformer, and the transformers + Bi-LSTM model, which reverses this order, both achieve better performance compared to Bi-LSTM + attention models in almost all metrics. Notwithstanding these commendable results, it is the Bi-LSTM and transformers concatenation model that outperforms all others, showing that concatenation is the most effective way of combining the unique capabilities of both the transformers' long-range dependency capturing and Bi-LSTM's local temporal dependency modeling strengths. While transformer-RNN hybrids certainly present promising improvements over earlier Bi-LSTM + attention models, the mode of integration is crucial to optimize their performance.

\subsubsection{Subtle Impact and Inferiority of Manual Feature Extraction} When evaluating the Bi-LSTM + attention model on both original smart meter data and manually feature-extracted data, a nuanced impact on the effectiveness of manual feature extraction emerges in the problem of residential occupancy detection. While the model trained on original data exhibits a marginally higher accuracy, precision, and ROC AUC, the recall and F1 score for the model trained on feature-extracted data are slightly higher. These metrics, along with the comparable distributions displayed by the box plots of the 10-fold cross-validation results, suggest a near-parity in performance between the two models using original and feature extracted data. 
% Despite certain metrics being slightly superior for one model over the other, the difference is minimal. 
This subtlety of variation indicates that manual feature extraction may not yield significant improvements when the data originates from diverse households. Consequently, relying on the innate feature extraction capabilities of neural networks could be an effective and more streamlined approach for occupancy detection using smart meter data.

\section{Conclusion}\label{sec:conclu}
This paper presented a compelling exploration of hybrid models combining both Bi-LSTM and transformer architectures for the task of residential occupancy detection using low-resolution smart meter data. Through effectively addressing complexities associated with traditional sensor-based methodologies and mitigating privacy concerns, this innovative approach utilizes deep learning techniques and underscores its suitability for large-scale deployments. By leveraging the strengths of Bi-LSTM and transformer in sequential and non-sequential processing and handling temporal dependencies locally and in a long range, our model achieves superior performance, evidenced by a range of evaluation metrics, including accuracy, precision, recall, F1, and ROC AUC. While these findings represent significant progress, they also shed light on potential future research directions, notably in the realm of unsupervised or semi-supervised occupancy detection for better use of unlabeled smart meter data.

\bibliographystyle{IEEEtran}
\bibliography{references}

\end{document}